\documentclass[conference]{IEEEtran}

\usepackage{url}            
\usepackage{booktabs}       
\usepackage{amsfonts}       
\usepackage{nicefrac}       
\usepackage{microtype}      
\usepackage{graphicx}
\usepackage{amsmath, xparse}
\usepackage{multirow}
\usepackage{hhline}
\usepackage{pifont} 
\newcommand{\xmark}{\ding{55}}
\usepackage{flushend}
\usepackage{balance}
\usepackage{subcaption}  

\IEEEoverridecommandlockouts
\usepackage{etoolbox}
\AtBeginEnvironment{equation}{\setlength{\abovedisplayskip}{3pt}}
\AtBeginEnvironment{equation}{\setlength{\belowdisplayskip}{3pt}}

\begin{document}
\bstctlcite{IEEEexample:BSTcontrol}
\title{Improving Robustness Against Adversarial Attacks with Deeply Quantized Neural Networks}

%

\author{\IEEEauthorblockN{Ferheen Ayaz\IEEEauthorrefmark{3},
Idris Zakariyya\IEEEauthorrefmark{1},
Jos\'e Cano\IEEEauthorrefmark{1}, \\
Sye Loong Keoh\IEEEauthorrefmark{1},
Jeremy Singer\IEEEauthorrefmark{1},
Danilo Pau\IEEEauthorrefmark{2},
Mounia Kharbouche-Harrari\IEEEauthorrefmark{2}}
\IEEEauthorblockA{\emph{
        \begin{tabular}{cc}
            \IEEEauthorrefmark{1}{University of Glasgow, UK} 
            \IEEEauthorrefmark{2}{STMicroelectronics}
            \IEEEauthorrefmark{3}{University of Sussex, UK}          
        \end{tabular}
   }}
\thanks{This work was done when Ferheen Ayaz was with University of Glasgow.}}

\maketitle 



\begin{abstract}

Reducing the memory footprint of Machine Learning (ML) models, particularly Deep Neural Networks (DNNs), is essential to enable their deployment into resource-constrained tiny devices. 
However, a disadvantage of DNN models is their vulnerability to adversarial attacks, as they can be fooled by adding slight perturbations to the inputs. 
Therefore, the challenge is how to create accurate, robust, and tiny DNN models deployable on resource-constrained embedded devices. 
This paper reports the results of devising a tiny DNN model, robust to adversarial black and white box attacks, trained with an automatic quantization-aware training framework, i.e. QKeras, with deep quantization loss accounted in the learning loop, thereby making the designed DNNs more accurate for deployment on tiny devices. 
We investigated how QKeras and an adversarial robustness technique, Jacobian Regularization (JR), can provide a co-optimization strategy by exploiting the DNN topology and the per layer JR approach to produce robust yet tiny deeply quantized DNN models. As a result, a new DNN model implementing this co-optimization strategy was conceived, developed and tested on three datasets containing both images and audio inputs, as well as compared its performance with existing benchmarks against various white-box and black-box attacks. 
Experimental results demonstrated that on average our proposed DNN model resulted in 8.3\% and 79.5\% higher accuracy than MLCommons/Tiny benchmarks in the presence of white-box and black-box attacks on the CIFAR-10 image dataset and a subset of the Google Speech Commands audio dataset respectively. It was also 6.5\% more accurate for black-box attacks on the SVHN image dataset.


\end{abstract}



\maketitle


\section{Introduction} 
\label{01_intro}

Deep Neural Networks (DNNs) demonstrate remarkable performance in various tasks such as natural language processing, cybersecurity, computer vision, intelligent applications and many more~\cite{liu2017survey}. However, DNN models are resource intensive with large memory footprint and computational requirements. Moreover, the increasing requirements of edge intelligence have given rise to new optimization strategies in Machine Learning (ML) which strive to reach optimal accuracy while shrinking the DNN model architectures at the same time~\cite{turner_iiswc_2018}.
The specific sub-discipline of ML that generates constrained ML workloads to be deployed on a target edge device, such as Microcontroller Units (MCUs) and sensors, is called Deeply Quantized Machine Learning (DQML). 
However, DNN models are often vulnerable to adversarial attacks causing changes to the input imperceptible to the human eye~\cite{attack1}. These vulnerabilities are critical, as they restrict the deployability of DNN models as an effective solution for real world applications such as autonomous cars, smart cities, intelligent applications and responsive Artificial Intelligence (AI)~\cite{lin2020adversarial}. Two widely used classes of adversarial attacks are \textit{white-box} and \textit{black-box} attacks. In white-box, the attacker has full knowledge of the DNN model, its structure and parameters, whereas in the black-box paradigm, the attacker is unaware of the used DNN model. In both scenarios, these attacks aim to cause deliberate mis-classifications or to disrupt the model performance. As such, there is an urgent need of balancing the trade-off between reducing the memory footprint of DNN models for tiny embedded devices while making them robust against adversarial attacks.

DQML offers a promise in terms on executing very low bit depth ML models on resource-constrained MCUs with $\mu$W power and a memory of only a few MBytes~\cite{TinyML2}. 
Various frameworks such as TensorFlow Lite (TFLite)~\cite{TFlite} for post-training optimization and Larq~\cite{Larq}, Brevitas~\cite{brevitas} and QKeras~\cite{Qkeras} for deep quantization-aware training, are used to optimize ML models resources. Most of these frameworks use quantization to optimize the utilized ML models based on data precision. 

QKeras is designed to offer quantization as low as a single-bit, and at the same time retaining the model accuracy through introducing quantization error in the form of random noise and learning to overcome it during training~\cite{Qkeras2}. It is based on drop-in replacement functions for Keras, thus providing the freedom to add a quantizer and choose quantization bit-depth separately for activations, biases and weights per layer. This is useful for efficient training of quantized DNN models. 
Among various deep quantization strategies offered by QKeras, there is stochastic quantization~\cite{Sto}, which instead of quantizing all elements (parameters) of a DNN model, quantizes a portion of the elements with a stochastic probability inversely proportional to the quantization error, while keeping the other portion unchanged in full-precision (FP). The quantized portion is gradually increased at each iteration until potentially the entire DNN is quantized. This procedure greatly and incrementally compensates the quantization error and thus yields better accuracy for very low-bit-depth DNN. 

In parallel, exploring the robustness of DNN models is critical to be integrated within DQML. In fact, enhancing the security while enabling the model's deployment in resource-constrained MCUs is a key challenge. 
Various defensive techniques and models for DNNs are present in the literature to provide robustness against adversarial attacks~\cite{silva2020opportunities}. However, such defensive mechanisms may result in an increased model size or accuracy drop for clean sets. Considering DQML models, which require extensive learning computation to reach optimal size and accuracy, with possible vulnerability to adversarial attacks, any addition to the model size or drop in accuracy can affect the deployment performance. In view of that, recent studies~\cite {{lin2019defensive,panda2020quanos}} have demonstrated that quantization can reduce computational requirements while granting robustness to a certain level of white-box adversarial attacks. 

Motivated by such an observation, this paper investigates the following hypothesis: \textit{Can a per-layer hybrid quantization scheme inherit robustness against white-box and black-box attacks while maintaining the trade-off between clean set accuracy performance and limited-resources requirements of tiny embedded devices?} We propose a DQML model that is deployable on tiny devices and highly robust to adversarial attacks, trained using the QKeras framework. Our theoretical investigation shows that QKeras utilizes Jacobian Regularization (JR) as an adversarial attack defensive mechanism. Based on this, we use QKeras to propose a Stochastic Ternary Quantized (STQ) DNN model with accuracy performance suitable for deployment in tiny MCUs, and potentially for image and audio in the same sensor package embodiment. Its ability to provide robustness against various adversarial attacks has been proven. The contributions of this paper are as follows:

\begin{itemize}
  \item Development of an STQ-based model that is less complex and can be deployed on MCUs with minimal memory footprint requirements and an improved accuracy performance on clean sets (Section~\ref{appr}).
  
  \item Analysis of the robustness of the STQ-based model according to its similarity with the JR technique, and demonstration of its resilience against various white and black box adversarial attacks (Sections~\ref{appr} and~\ref{eval}).
  
  \item A comprehensive comparison of the resilience of the STQ-based model to industry endorsed MLCommons/Tiny benchmarks for both image and audio inputs while investigating its performance with $K$-fold cross validation (Section~\ref{eval}).
\end{itemize}

\section{Background and Related Work} 
\label{backrw}


\subsection{Adversarial Attacks Against Machine Learning Models}

Security considerations are important for sensitive applications like intelligent transportation systems, healthcare and financial systems which utilize image and voice recognition AI-based models~\cite{sikder2021survey}. A disadvantage of ML and AI models developed for such tasks is that they can be vulnerable to attacks which can compromise their integrity, confidentiality and privacy in real world applications. 
In particular, adversarial attacks involve adding a small perturbation to the input to maximize the loss function of a model under a constrainted norm~\cite{su2019one}. 
Equation (\ref{eqadp}) expresses such a procedure of introducing a perturbation into an input data, where: $\mathcal{L}(\theta, x', y)$ is the loss function, $\theta$ denotes the model parameters, $x'$ is the perturbed input, $y$ is the model output, $\delta$ denotes the perturbation and $p$ is the perturbation norm~\cite{jiang2021project}. This work considers the two types of attacks defined in Section~\ref{01_intro}, i.e. white-box and black-box attacks.

A common adversarial attack technique is the Fast Gradient Sign Method (FGSM)~\cite{huang2020bridging}. This is a white-box attack which uses a single-step iteration to estimate the gradient of the model training loss function based on the inputs. An FGSM~\cite{huang2020bridging} attack procedure is expressed in (\ref{fgsm}), where $\Delta$ represents the gradient and $\epsilon$ denotes a small constant value that restricts the perturbation. A variation of FGSM is the Projected Gradient Descent (PGD)~\cite{huang2020bridging}. This is a more computationally expensive multi-step threat model which runs several iterations to find an adversarial input with the lowest possible $\delta$, as expressed in (\ref{pgd}), where $i$ is the iteration index, $\alpha$ denotes the gradient step size and $S$ represents the perturbation set. 
\vspace{-1pt}
\begin{equation}
\label{eqadp}
\smash{\displaystyle\max_{|| \delta||_p}}\, \mathcal{L}(\theta, x', y)   
\end{equation} 
\vspace{-4pt}
\begin{equation}
\label{fgsm}
    x' = x + \epsilon \cdot sign(\Delta_x \mathcal{L}(\theta, x, y))
\end{equation}
\noindent \vspace{-4pt}
\begin{equation}
\label{pgd}
    x_{i+1}' = \prod_{x+S} (x + \alpha \cdot sign(\Delta_x \mathcal{L}(\theta, x, y)))
\end{equation}

An efficient black-box attack is the Square attack~\cite{square}, which is based on a random search optimization technique with multiple iterations. In each iteration, it changes a small fraction of the input shaped into square at random positions. Similar to gradient-based optimizations, it also relies on step-size reduction, where the size refers to the dimensions of the square~\cite{square}. 
Another form of black-box attack is the Boundary attack~\cite{boundary}, where the queries are used to estimate the decision boundaries of the output classes. Starting with a clean image, gradient estimation is performed with queries, moving along the estimated direction in each iteration and projecting a new perturbation until the model decision is changed~\cite{boundary}. 

Overcoming the white-box and black-box attack methods requires a suitable defensive mechanism. Therefore, it is important to enhance the model performance against different attacks variations to ease its deployment.


\begin{figure*} 
	\centering
	\includegraphics[scale=0.40]{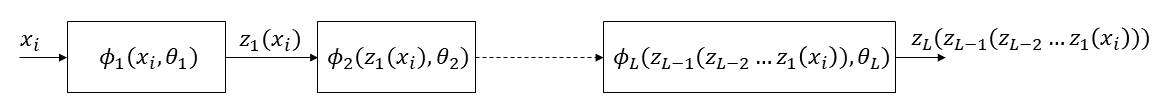}
	\caption{Transformation of input $x_i$ into output $z_L$ by DNN.}
	\label{fig2}
\end{figure*}

\subsection{Robustness against Adversarial Attacks}

Various defense methods have been proposed to increase the robustness of DNNs against adversarial attacks~\cite{xu2020adversarial}. Some strategies aim at detecting adversarial inputs~\cite{detect} or performing transformations to remove perturbations~\cite{transform}~\cite{Bernhard_LuringAE_IJCNN21} through an additional network or module. 
Adversarial training introduces adversarial inputs during the model's learning so that it learns not to misinterpret them~\cite{AT}. However, these approaches do not guarantee robustness against attacks which are not introduced to the model during training, as they are not learned by the model. Other methods such as~\cite{picot2022adversarial} decrease the model's sensitivity to small perturbations by adding a regularization term in the loss function. Equation (\ref{joinl}) expresses this joint loss function, where $\mathcal{L}_{reg}$ denotes the regularization term and $\lambda$ is a hyper-parameter used to allow the adjustment between the regularization and the actual loss.  
\begin{equation}
\label{joinl}
\mathcal{L}_{joint} = \mathcal{L}(\theta, x, y) + \mathcal \lambda{L}_{reg},  
\end{equation}


\subsection{Jacobian Regularization}

Jacobian Regularization (JR) is a technique to provide adversarial robustness, where the 
input gradient regularization normalizes the gradient of the cross-entropy loss, such that  
$\mathcal{L}_{reg} = \sum_{k=1}^K|| \Delta_x z_k^L (x_i) ||_2^2 = ||J(\mathbf{x}_i)||_F^2$ and $||J(\mathbf{x}_i)||_F^2$ is the Frobenius norm of the model’s Jacobian matrix evaluated on the input data~\cite{Jacobian1}. To reduce computational complexity, per-layer JR is proposed in~\cite{Jacobian2}. 

To demonstrate the robustness of per-layer JR, let's consider a $D$-dimensional network input $X$ consisting of $N$ training samples. As shown in Figure~\ref{fig2}, the network contains $l=1,2,...,L$ layers. $z_l$ denotes the output at layer $l$ and $z_l^k$ is the output of the $k^{th}$ neuron of the $l^{th}$ layer. Consider the softmax operation at the output of the network; the predicted final output for computing the top-1 accuracy for an input $x_i$ is $f(x_i)=argmax \{ z_L^1, z_L^2,..,z_L^K \}$, where $K$ is the dimension of the output vector. 
The Jacobian matrix of a DNN is computed at $L^{th}$ layer, i.e. $J_L(x_i) = \nabla_x z_L (x_i)$, and is defined as:
\begin{equation}
	J_L (x_i) = 
	\begin{bmatrix}
		\frac{ \partial z_L^1 x_1 }{ \partial x_1} & \frac{ \partial z_L^1 x_1 }{ \partial x_2} & ... & \frac{ \partial z_L^1 x_1 }{ \partial x_D} \\
		\frac{ \partial z_L^2 x_2 }{ \partial x_1} & \frac{ \partial z_L^2 x_2 }{ \partial x_2} & ... & \frac{ \partial z_L^2 x_2 }{ \partial x_D} \\
		. & . & . & . \\
		. & . & \;\; . & . \\
		\frac{ \partial z_L^K x_K }{ \partial x_1} & \frac{ \partial z_L^K x_K }{ \partial x_2} & ... & \frac{ \partial z_L^K x_K }{ \partial x_D} \\
			
	\end{bmatrix} \epsilon \mathbb{R} ^{K \times D}.
\end{equation}

In~\cite{Jacobian1}, the JR term for an input $x_i$ is defined as
\begin{equation}
\small
	||J(x_i)||_F^2 = \sum_{d=1}^D \sum_{k=1}^K \Big( \frac {\partial} {\partial x_d} z_L^k (x_i) \Big)^2 = \sum_{k=1}^K ||\nabla_x z_L^k (x_i) ||^2_2,
\label{Jreg}
\end{equation}
The standard loss of the training is added with the regularization term in (\ref{Jreg}) to improve the robustness of the DNN. It is proposed as a post-training, in which the network is re-trained for fewer iterations with the new loss function~\cite{Jacobian1}. As $i \epsilon \{1,2,...,N\}$, JR requires the computation of $N$ gradients, whereas in per-layer JR, the Jacobian matrix is computed on only one random $i$ at each layer~\cite{Jacobian2}. The basic idea is to reduce the Frobenius norm of the Jacobian matrix which results in the expansion of the classification margin, i.e. the distance between an input and the decision boundary induced by a network classifier.


\subsection{Adversarial Robustness of Quantized Models}

Model quantization techniques are widely used in various fields~\cite{gholami2021survey}. In~\cite{online2}, it is used for anomaly detection and thwarting cyber attacks in Internet-of-Things (IoT) networks. 
However, the limitation of a quantized model is the shift of the FP model classification boundary, which may influence how vulnerable the model is to adversarial perturbations~\cite{song2020improving}. As such, the authors in~\cite{song2020improving} investigated the use of a boundary-based retraining method to reduce adversarial and quantization losses with the usage of non-linear mapping as a defensive mechanism against white-box adversarial attacks. 
Other previous studies explored the impact of perturbations on different models. Table~\ref{mlcompar} compares some existing works that investigated the robustness of quantized DNN models. These studies investigated the resilience of quantized DNN models without examining their deployment feasibility on resource-constrained devices, such as MCUs or sensors, and most of them considered only white-box attacks.

\begin{table}[!tp]
\caption{Comparison of adversarially quantized DNN models.}
\centering
\fontsize{6.5}{9}\selectfont
\label{mlcompar}
\begin{tabular}{|cc|cccc|c|}
\hline
\multicolumn{2}{|c|}{\textbf{Works}} &
  \multicolumn{1}{c|}{\cite{galloway2017attacking}} &
  \multicolumn{1}{c|}{\cite{bernhard2019impact}} &
  \multicolumn{1}{c|}{\cite{kim2020robust}} &
  \cite{gorsline2021adversarial} &
  Ours \\ \hline
\multicolumn{1}{|c|}{\multirow{3}{*}{\begin{tabular}[c]{@{}c@{}}\textbf{Compress.} \\ \textbf{technique}\end{tabular}}} &
  Method &
  \multicolumn{4}{c|}{N bit-width quantization} &
  SQ* \\ \cline{2-7} 
\multicolumn{1}{|c|}{} &
  on weights &
  \multicolumn{1}{c|}{\checkmark} &
  \multicolumn{1}{c|}{\checkmark} &
  \multicolumn{1}{c|}{\checkmark} &
  \checkmark &
  \checkmark \\ \cline{2-7} 
\multicolumn{1}{|c|}{} &
  on activat. &
  \multicolumn{1}{c|}{\checkmark} &
  \multicolumn{1}{c|}{\checkmark} &
  \multicolumn{1}{c|}{\checkmark} &
  \xmark &
  \checkmark \\ \hline
\multicolumn{1}{|c|}{\multirow{2}{*}{\textbf{Inputs}}} &
  Images &
  \multicolumn{1}{c|}{\checkmark} &
  \multicolumn{1}{c|}{\checkmark} &
  \multicolumn{1}{c|}{\checkmark} &
  \checkmark &
  \checkmark \\ \cline{2-7} 
\multicolumn{1}{|c|}{} &
  Audio &
  \multicolumn{1}{c|}{\xmark} &
  \multicolumn{1}{c|}{\xmark} &
  \multicolumn{1}{c|}{\xmark} &
  \xmark &
  \checkmark \\ \hline
\multicolumn{2}{|c|}{\textbf{Datasets}} &
  \multicolumn{1}{c|}{\begin{tabular}[c]{@{}c@{}}MNIST\\ C-10*\end{tabular}} &
  \multicolumn{1}{c|}{\begin{tabular}[c]{@{}c@{}}MNIST\\ SVHN\end{tabular}} &
  \multicolumn{1}{c|}{\begin{tabular}[c]{@{}c@{}}MNIST\\ C-10*\\ TinyI*\end{tabular}} &
  \begin{tabular}[c]{@{}c@{}}MNIST\\ Spiral\end{tabular} &
  \begin{tabular}[c]{@{}c@{}}C-10*\\ SVHN\\ GSC\end{tabular} \\ \hline
\multicolumn{1}{|c|}{\multirow{6}{*}{\textbf{Attacks}}} &
  \begin{tabular}[c]{@{}c@{}}White\\ box\end{tabular} &
  \multicolumn{1}{c|}{\begin{tabular}[c]{@{}c@{}}FGSM\\ PGD\\ C\&W\end{tabular}} &
  \multicolumn{1}{c|}{\begin{tabular}[c]{@{}c@{}}FGSM\\ BIM\\ C\&W\end{tabular}} &
  \multicolumn{1}{c|}{\begin{tabular}[c]{@{}c@{}} \xmark \end{tabular}} &
  FGSM &
  \begin{tabular}[c]{@{}c@{}}FGSM\\ PGD\\ C\&W\end{tabular} \\ \cline{2-7} 
\multicolumn{1}{|c|}{} &
  \begin{tabular}[c]{@{}c@{}}Black\\ box\end{tabular} &
  \multicolumn{1}{c|}{\xmark} &
  \multicolumn{1}{c|}{\begin{tabular}[c]{@{}c@{}}SPSA \\ ZOO \end{tabular}} &
  \multicolumn{1}{c|}{\xmark} &
  \xmark &
  \begin{tabular}[c]{@{}c@{}}Square\\ Boundary\\ ZOO\end{tabular} \\ \cline{2-7}
\multicolumn{1}{|c|}{} &
  Random &
  \multicolumn{1}{c|}{\xmark} &
  \multicolumn{1}{c|}{\xmark} &
  \multicolumn{1}{c|}{\begin{tabular}[c]{@{}c@{}}Gaus.\\ noise\end{tabular}} &
  \xmark & \xmark
   \\ \hline
  
\multicolumn{2}{|c|}{\textbf{Memory footprint}} &
  \multicolumn{1}{c|}{\xmark} & \multicolumn{1}{c|}{\xmark} & \multicolumn{1}{c|}{\xmark} & \xmark & 410 KB \\ \hline
\multicolumn{7}{l}{*C-10 denotes CIFAR-10 dataset, *TinyI: TinyImageNet, *SQ: Stochastic Quantization}
\end{tabular}
\end{table}

\section{Proposed DNN model} 
\label{appr}

This section introduces a new robust and less complex DNN model based on the Stochastic Ternary Quantization (STQ) approach which is conceived to be deployable on tiny MCUs.


\subsection{Adversarial Robustness in QKeras}

\begin{figure}[tp]
  \centering
  \includegraphics[width = 1.0\columnwidth]{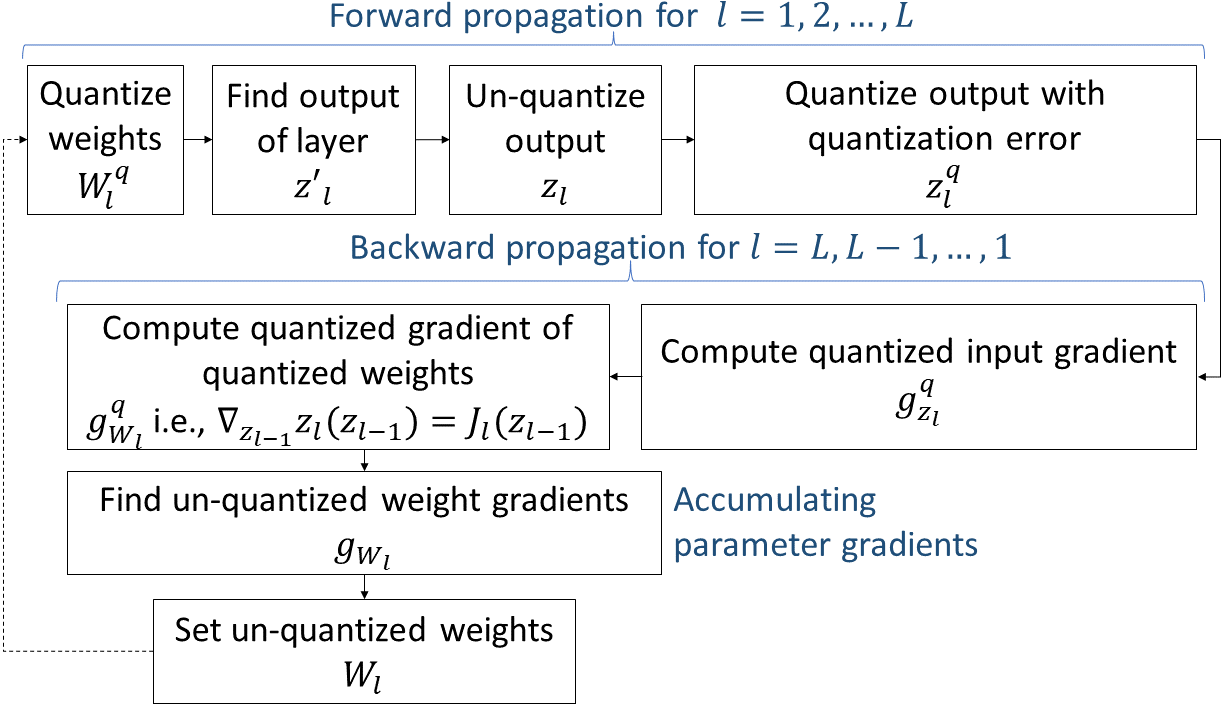}
   \caption{QKeras quantization aware training flow.}
 	\label{fig1}
\end{figure}

QKeras~\cite{Qkeras} is a DNN framework targeting quantization based on Keras~\cite{keras},
which provides a productive methodology to build and train quantized neural networks, either fractional or integer spanning from 1 to 32 bits. QKeras performs quantization aware training~\cite{qkeras3}. The background algorithm to train neural networks with $q$ bit-width weights, activations and gradient parameters is conceptualized in~\cite{dorefa}. 
In particular, the network training in QKeras includes a backward propagation where parameter gradients are stochastically quantized into low bit-width numbers. Figure~\ref{fig1} shows the process flow of training each layer in QKeras. Each training iteration involves a forward propagation step to quantize weights, find output and add quantization error into the output of each layer. Then it includes a backward propagation step where gradients are stochastically quantized into low bit-width numbers. Finally, unquantized weights and gradients are updated for the next iteration. The training process of deeply quantized networks, through QKeras, make them robust to adversarial attacks due to the two following reasons: i) a noise function is introduced during quantization of gradients to overcome quantization error during training, which makes a DNN robust to noise and perturbation effect; ii) QKeras involves JR features that are explained below.

\textbf{Theorem:} \textit{QKeras introduces per-layer JR and therefore increases the robustness of DNNs against adversarial attacks.}

\textbf{Proof:} Considering the per-layer structure of DNNs, as shown in Figure~\ref{fig2}, the output $z_L$ at the last layer $L$ is 
\begin{equation}
	z_L = \phi_L((\phi_{L-1}(...\phi_1(x_i, \theta_1),...\theta_{L-1}), \theta_L)),
\end{equation}
where $\phi_l(., \theta_l)$ represents the function of the $l^{th}$ layer, $\theta_l$ denotes the model parameters at layer $l$, and $z_0 = x_i$~\cite{Jacobian2}. The Jacobian matrix of the $l^{th}$ layer is defined as 
\begin{equation}
J_l(z_{l-1}) = \frac{dz_l}{dz_{l-1}},
\label{back}
\end{equation}
which is back-propagated during QKeras training (Step 10-16 of Algorithm 1 in~\cite{dorefa}). The derivation expressed in (\ref{back}) is the Jacobian matrix~\cite{Jacobian2} of the $l^{th}$ block layers. Thus, QKeras with such patterns incorporates per-layer JR during training.

To prove that per-layer JR enhances adversarial robustness, consider a clean input $x_c$ and an adversarial input $x_p$, both close to an input $x_i$ and all belonging to the same class $k$. Since $f(x_i) = f(x_c) \neq f(x_p)$, then the $\ell_2$ distance metric of the input and output of the network, as defined in~\cite{Jacobian1}, is
\begin{equation}
	\frac{||x_p - x_i||_2}{||x_c - x_i ||_2}  \approx 1,
\label{p1}
\end{equation}

\begin{equation}
	\frac{||z_L (x_p) - z_L (x_i)||_2}{||z_L (x_c) - z_L (x_i) ||_2}  > 1,
\label{p2}
\end{equation}
Combining (\ref{p1}) and (\ref{p2}) and using the Mean Value Theorem~\cite{shishkina2022mean}, it is justified that a lower Frobenius norm makes a network less sensitive to perturbations, i.e.,
\begin{equation}
\frac{||z_L (x_p) - z_L (x_i)||_2^2}{||x_p - x_i ||_2^2}  \leq ||J(x')||_F^2,
\end{equation}
where $x' \epsilon [x_i, x_p]$. Similar to (\ref{p2}), for each layer $l$, we have
\begin{equation}
	\frac{||z_l (x_p) - z_l (x_i)||_2}{||z_l (x_c) - z_l (x_i) ||_2}  > 1.
	\label{p3}
\end{equation}
The misclassification error is propagated at each layer, thus
\begin{equation}
	\frac{||z_l (z_{l-1}(x_p)) - z_l (z_{l-1}(x_i))||_2}{||z_l (z_{l-1}(x_c)) - z_l (z_{l-1}(x_i)) ||_2}  > 1.
	\label{p4}
\end{equation}

This error is back-propagated and then adjusted at each layer. Therefore the learning optimization process increases robustness of a DNN model trained by QKeras since it discriminates the error due to the clean versus the perturbed input. Consequently, it can be concluded that QKeras deep quantization-aware process introduces per-layer JR with respect to the previous layer’s input that is back propagated, as shown in Figure~\ref{fig1} and Step 11 to 15 of Algorithm 1 in~\cite{dorefa}. Although back propagation is intended to quantize parameters, it ultimately results in more robust networks. Its learning  process increases the training computation time but its advantage is two-fold, i.e. deep quantization and robustness to adversarial perturbations.


\subsection{Stochastic Ternary Quantized QKeras Architecture}

\begin{figure*} [!tp]
  \centering
  \includegraphics[width=0.95\textwidth]{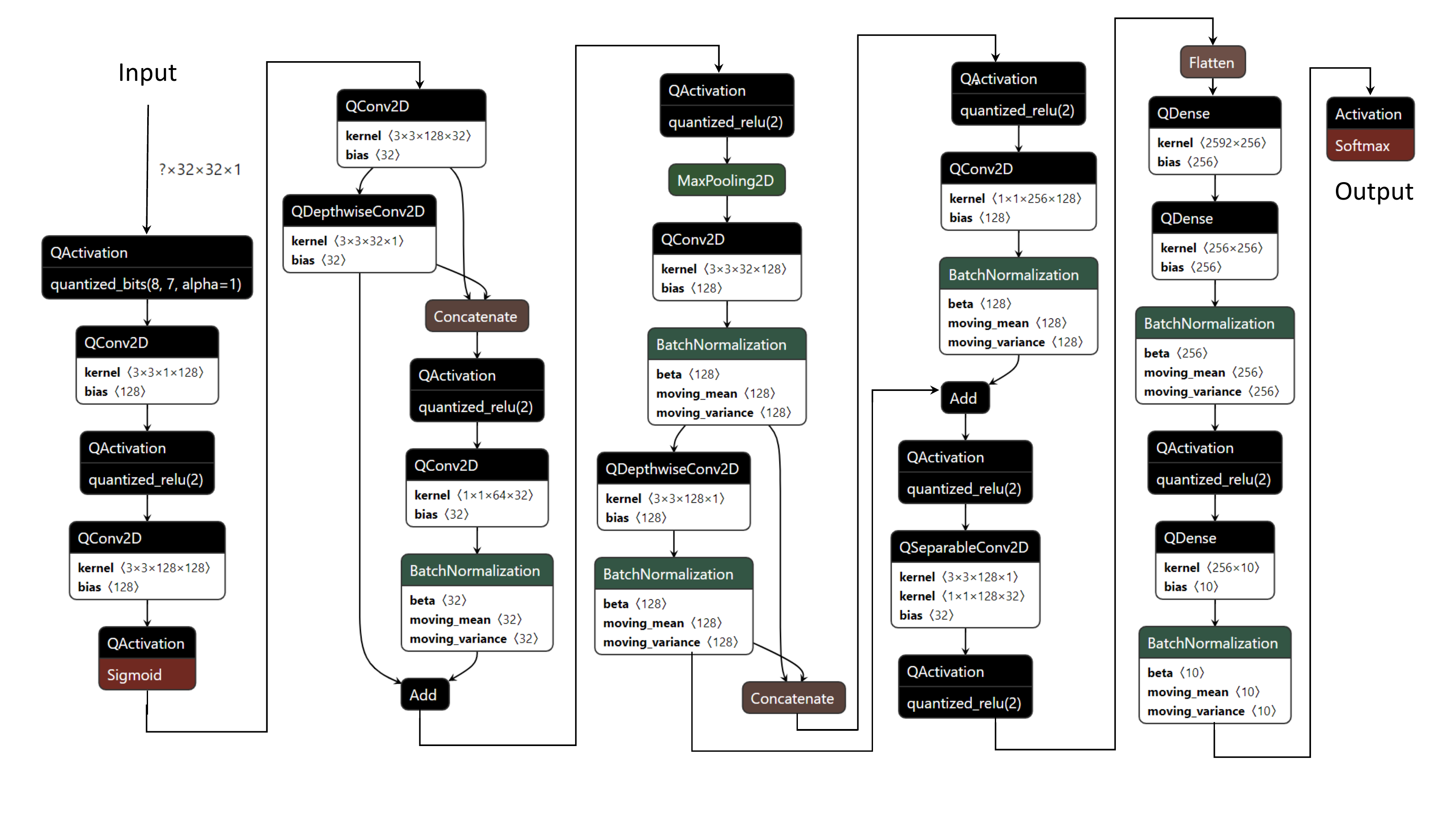}
  \vspace{-0.5cm}
   \caption{Proposed architecture of the STQ-based DNN model developed using QKeras.}
 	\label{fig3}
\end{figure*}

As the QKeras~\cite{qkeras3} API serves as an extension of Keras, initially a 32-bit FP DNN model was built with~\cite{keras}. The FP model consists of six convolutional layers including depth-wise, separable, and three fully connected layers. The former layers are used for capturing channel-wise correlations and can provide more features with less parameters, particularly with image inputs~\cite{guo2019depthwise}. 
A multi-branch topology was used with residual connections to refine the feature maps. For better accuracy, batch normalization~\cite{wang2021enabling}, ReLU~\cite{eckle2019comparison} and sigmoid~\cite{papernot2021tempered} activation functions are considered in the hidden layers, while softmax is used in the output layer. 
The FP model contains 998,824 parameters, of which 997,460 are trainable and 1,364 are non-trainable, as returned by \texttt{model.summary()}. This DNN model is not integrated with JR by default, and is therefore vulnerable to adversarial attacks. However, the model cannot be deployed on tiny MCUs due to its large size (4~MB). Since our target device is the STM32H7, which has a maximal RAM memory of 1~MB, we applied the STQ method to this model using QKeras.

Figure~\ref{fig3} shows the architecture of the STQ model visualized using Netron~\cite{netron} to render the DNN graph along with its hyperparameters. As we can see, the convolutional, depth-wise, separable layers and activation are appended with \textit{Q}, which indicates the quantization version of the FP Keras~\cite{keras} model, while the \textit{quantized\_relu} represents the quantized activation function versions of Keras. The proposed STQ model used the heterogeneous quantization features of QKeras, which supports independent quantization of each layer in the DNN~\cite{Qkeras2}. This is useful in reducing the model memory footprint and complexity with increased performance accuracy.

\section{Evaluation} 
\label{eval}

This section describes the evaluation procedure of the proposed STQ model, using image and audio datasets for clean and adversarial samples. Moreover, it compares the top-1 accuracy of our STQ model with other benchmarks for three black-box and three white-box adversarial attacks.


\subsection{Experimental Setup}

\subsubsection{Datasets and Pre-processing}

To evaluate the effectiveness of the devised STQ model, the following audio and image benchmark datasets are considered:

\begin{itemize}
  \item \textbf{CIFAR-10} consists of 60,000 images belonging to 10 different classes. Each class is divided into 50,000 training images and 10,000 test images~\cite{cifar10}.
  
  \item \textbf{Street View House Numbers (SVHN)} is a real-world image dataset obtained from house numbers in Google Street View images. It consists of 10 classes, one for each digit. There are 73,257 and 26,032 digits for training and testing respectively~\cite{SVHN}.
  
  \item \textbf{Google Speech Commands (GSC)} consists of 65,000 one-second long utterances of 30 short words by thousands of different people. Its 12 classes comprise words of `yes', `no' and digits from `zero' to `nine' that were used in the experiment. The number of training and test samples are 31,257 and 15,636 respectively~\cite{GSC}.
\end{itemize} 

Note that the input image samples were normalized between 0 and 1 pixel and then converted into gray-scale images before presenting them to the DNN input. This effectively reduced the computational cost of processing the image samples while avoiding the use color which are known to be deceitful. Color processing was out of scope by this work and may be considered in future extensions of it.


\subsubsection{Model Training Procedure}

Table~\ref{tableparam} lists the training parameters used to evaluate the full-precision (FP), the proposed STQ and other tested quantized models. A cosine annealing Learning Rate (LR) function~\cite{loshchilov2016sgdr} was used with the Adamax optimizer for faster convergence. These training parameters were selected to both fine tune each model and reduce its computational complexity, while maintaining better or state-of-the-art performance.

\begin{table}[!ht]
\caption{Training parameters for QKeras DNN}
  \label{tableparam}
  \centering
  \begin{tabular}{|c|c|}
  \hline
  \textbf{Parameter} & \textbf{Value} \\ \hline
  {Epochs} & {1000} \\ \hline
  {Batch Size} & {64} \\ \hline 
  {Learning Rate} &  [1 $\times$ $10^{-6}$, 1 $\times$ $10^{-3}$] \\ \hline
  {LR Scheduler} & {Cosine Annealing} \\ \hline
  {Loss} & {Categorical Cross-entropy} \\ \hline
  {Optimizer} & {Adamax} \\ \hline
 
  \end{tabular}
\end{table}

Since the proposed STQ model targets images and audio datasets, ResNetv1 and DS-CNN TFLite models were also used and tested for comparison purposes. This is to investigate the robustness of the model against industry adopted MLCommons/Tiny models, which can provide insights into the performance capability of the proposed model across various datasets and other benchmarks.


\subsubsection{Adversarial Attacks Procedure}

The proposed STQ model was evaluated against several adversarial attacks to demonstrate its resilience and robustness. Three white-box attacks, FGSM and PGD~\cite{huang2020bridging} and Carlini and Wagner (C\&W-L2)~\cite{carlini2017towards}; and three black-box attacks, Square~\cite{square}, Boundary attack~\cite{boundary} and Zero Order Optimization (Zoo)~\cite{chen2017zoo} were considered. 
The perturbed data samples for all attacks were generated with the Adversarial Robustness Toolbox (ART)~\cite{nicolae2018adversarial} against the tested datasets. For FGSM and PGD samples, an $\epsilon$ value of 0.6 with an L1 norm was used. For the Square attack, an $\epsilon$ value of 0.6 with \emph{infinity} norm was used, while for the Boundary attack, an $\epsilon$ value of 0.01 with \emph{infinity} norm was used. The maximum number of iterations used for C\&W-L2 and Zoo is 10, and a binary search tree of 10 used for Zoo. The C\&W-L2 and Zoo are used with CIFAR-10 data samples. The number of samples used to create the perturbations were 100, 240 and 2400 for CIFAR-10, SVHN and GSC datasets respectively. 

To further examine the strength of our STQ model against FGSM, PGD~\cite{huang2020bridging} and Square~\cite{square} attacks, they were crafted under different attack strengths using the 10,000 test samples of CIFAR-10. The FGSM attacks were crafted for $\epsilon$ ranging in: $\{0.05, 0.1, 0.15, 0.2, 0.25, 0.3\}$ as denoted in~\cite{panda2020quanos}. Regarding the PGD attack, an iteration $t =7$ was used along with a step size $\alpha = 2/255$ and an $\epsilon$ ranging in: $\{8/255, 16/255, 32/255\}$. For the Square~\cite{square} attacks, the first variation consists of \emph{infinity} norm, $\epsilon$ value of 0.05 and maximum iterations of 10,000. The second variation of the square attack uses the $\ell_2$ norm and maximum iterations of 10,000 as denoted in \cite{qin2021random}. 


\subsubsection{$K$-fold Cross Validation}

In order to estimate the variance of the clean and adversarial data samples across each tested model, $K$-fold cross validation was used. This technique splits the entire dataset into $K$ (fold) equal subsets, of which $K-1$ subsets are used for training and the remaining one subset is used for validation~\cite{berrar2019cross}. For each fold, the model is fitted on the training set and predicted on the validation set to estimate the average performance. For implementation purposes, $K=10$ was considered, because a larger number of folds may increase the predictive performance~\cite{wong2019reliable}. In addition, the scikit-learn~\cite{sklearn} ML Python API was used.


\subsection{Quantization schemes}

Table~\ref{table:QKeraseval} shows the performance (top-1 test accuracy) comparison between different quantized models against the tested datasets using clean inputs from each dataset. 
In addition, it includes the FLASH MCU memory of each quantized model computed based on the weight profile obtained using the \texttt{qstats} QKeras library. 
As we can see, the STQ-based model provides the highest accuracy as compared to the FP, Stochastic Binary (S-Binary), and other quantized models. Moreover, this model is lighter than the FP, 8-bit and 4-bit models and can be deployed on MCU devices. These results motivate further investigations on the robustness of the proposed STQ model against adversarial attacks, since the FP model does not have any integrated defensive mechanism, nor better accuracy performance in classifying clean samples.

\begin{table}[!t]
\caption{Performance (top-1 accuracy) comparison on CIFAR-10, SVHN and GSC clean datasets.}
\begin{center}
\label{table:QKeraseval}
\begin{tabular}{|c|c|c|c|c|c|c|}
\hline
\textbf{\multirow{2}{*}{Model}} & \textbf{Flash} & \textbf{CIFAR-10}  & \textbf{SVHN} & \textbf{GSC} \\
& \textbf{(KB)} & \textbf{Acc. (\%)}  & \textbf{Acc. (\%)}  & \textbf{Acc. (\%)}          \\ \hline
{FP} & {4496} & {52.73} &{67.94} &{89.50}   \\ \hline
{8-bit}   & {1124} & {52.46} &{70.04} &{85.60}   \\ \hline
{4-bit}  & {527} & {50.88} &{64.72} &{82.06}   \\ \hline
{Ternary}  & {410} & {77.35} &{93.11} &{88.30} \\ \hline
{STQ} & {410} & \textbf{80.57} &\textbf{95.53} &\textbf{94.76} \\ \hline
{2-bit} & {281} & {46.35} &{89.74} &{81.56}   \\ \hline
{Binary}   & {140} & {39.03} &{56.91} &{77.10} \\ \hline
{S-Binary} & {140} & {52.25} &{63.72} &{82.51}\\
\hline
\end{tabular}
\end{center}
\end{table}


\subsection{Robustness evaluation against adversarial attacks}

\begin{figure}
    \centering
    \centering {{\includegraphics[width=0.30\textwidth]{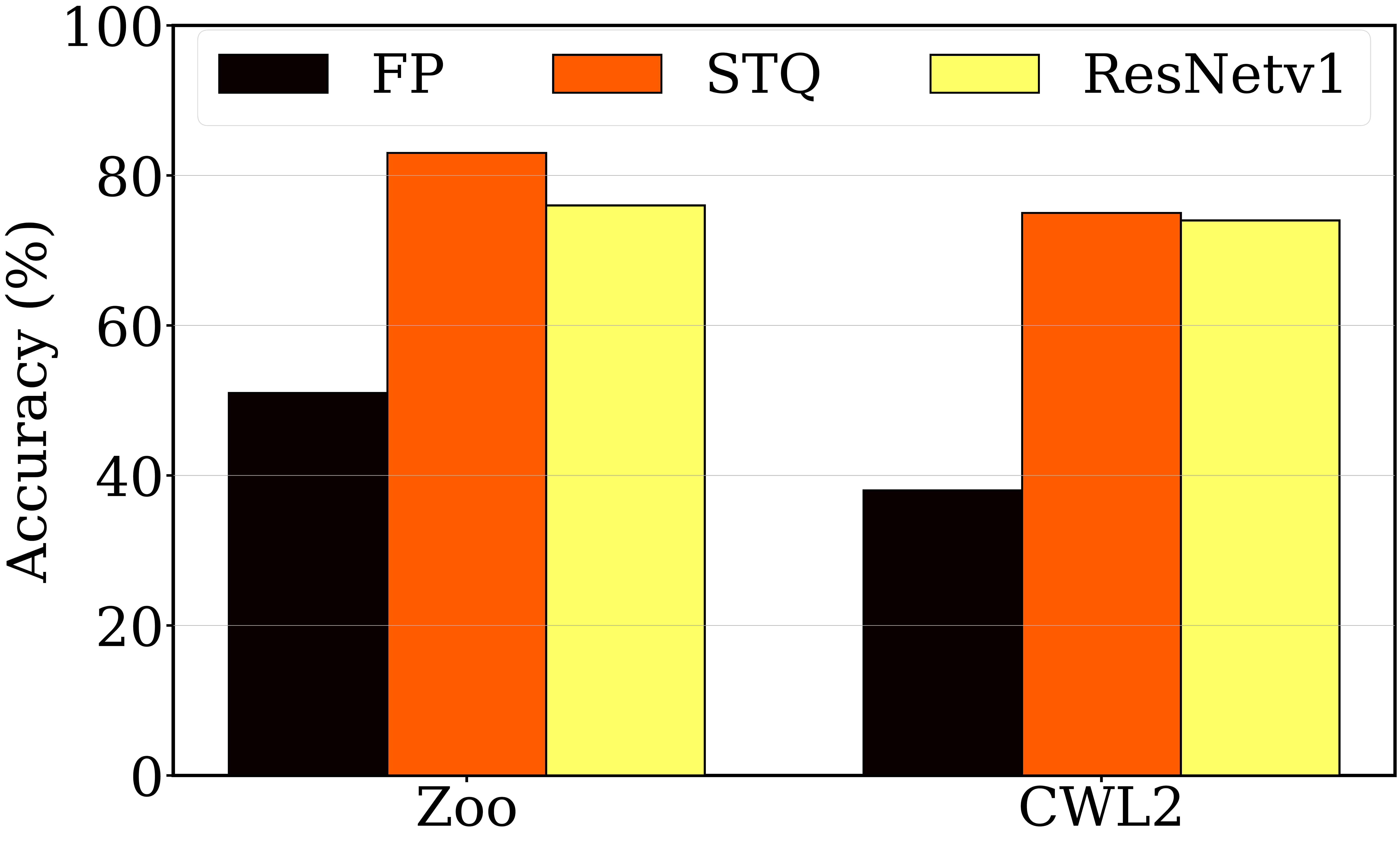}}}
    \caption{Models robustness (top-1 test accuracy) comparison against Zoo and C\&W-L2 attacks for the CIFAR-10 dataset.}
    \label{fgsmrcomp}
\end{figure}

Table~\ref{table:modrcomp} shows the robustness (top-1 test accuracy) of our STQ model against two white-box attacks (FGSM, PGD) and two black-box attacks (Square, Boundary). The results are compared with two MLCommons/Tiny benchmarks: ResNetv1 for image classifications trained on CIFAR-10 and SVHN datasets; and DS-CNN TFLite for Keyword Spotting trained on GSC dataset~\cite{MLperf}. 
In addition, Figure~\ref{fgsmrcomp} shows the robustness against another white-box attack (C\&W-L2) and a black-box attack (Zoo). The results are compared with ResNetv1 for image classifications trained on the CIFAR-10 dataset.

\begin{table}[tp] 
\caption{Models robustness (top-1 test accuracy) comparison for CIFAR-10, SVHN, and GSC datasets.}
\begin{center}
\fontsize{6.5}{9}\selectfont
\label{table:modrcomp}
\begin{tabular}{|c|c|c|c|c|c|c|c|}
\hline
\textbf{Dataset}& \textbf{Model} & \textbf{Clean } & \textbf{FGSM} & \textbf{PGD} & \textbf{Square} & \textbf{Boundary} \\ 
&& \textbf{Acc(\%)} &  \textbf{Acc(\%)} & \textbf{Acc(\%)} &  \textbf{Acc(\%)} & \textbf{Acc(\%)}  \\ \hline
\multirow{3}{*}{CIFAR-10} &{ResNetv1} & {85.0} & {65.6} & {64.1} & {65.5} & {82.9}    \\ \hhline{~-||-----} 
&{STQ} & {80.6} & \textbf{80.8} & \textbf{80.8} & \textbf{73.6} & \textbf{83.8} \\ \hline 
\multirow{3}{*}{SVHN} &{ResNetv1} & {94.8} & \textbf{94.0} & \textbf{94.0} & {82.1} & {94.0}    \\ \hhline{~-||-----} 
&{STQ} & \textbf{95.5} & {93.1} & {93.1} & \textbf{91.7} & \textbf{97.3} \\ \hline
\multirow{3}{*}{GSC} &{DS-CNN} & {89.1} & {15.0} & {15.4} & {15.1} & {14.7}    \\ \hhline{~-||-----} 
&{STQ} & \textbf{94.8} & \textbf{94.3} & \textbf{94.2} & \textbf{94.3} & \textbf{95.4} \\ \hline
\end{tabular}
\end{center}
\end{table}

As we can see in Table~\ref{table:modrcomp}, there is a drop in accuracy for the two MLCommons/Tiny benchmarks against all adversarial attacks, as no defense mechanism is included, which makes STQ with integrated JR an interesting scheme. 
On average, adversarial attacks have caused an accuracy drop of 15.5\%, 3.8\% and a massive 74.1\% with CIFAR-10, SVHN and GSC datasets respectively. 
Related to the C\&W-L2 and Zoo attacks, we can see in Figure~\ref{fgsmrcomp} that the average drop of ResNetv1 for the CIFAR-10 dataset is 10.5\%. All the previous accuracy drops highlight the importance of robust tiny models. Note that Figure~\ref{fgsmrcomp} also includes the robustness of our FP model, which is clearly lower than STQ.

\begin{figure*} 
\centering
  \begin{subfigure}[b]{0.32\linewidth}
    \centering
    \includegraphics[width=0.99\linewidth]{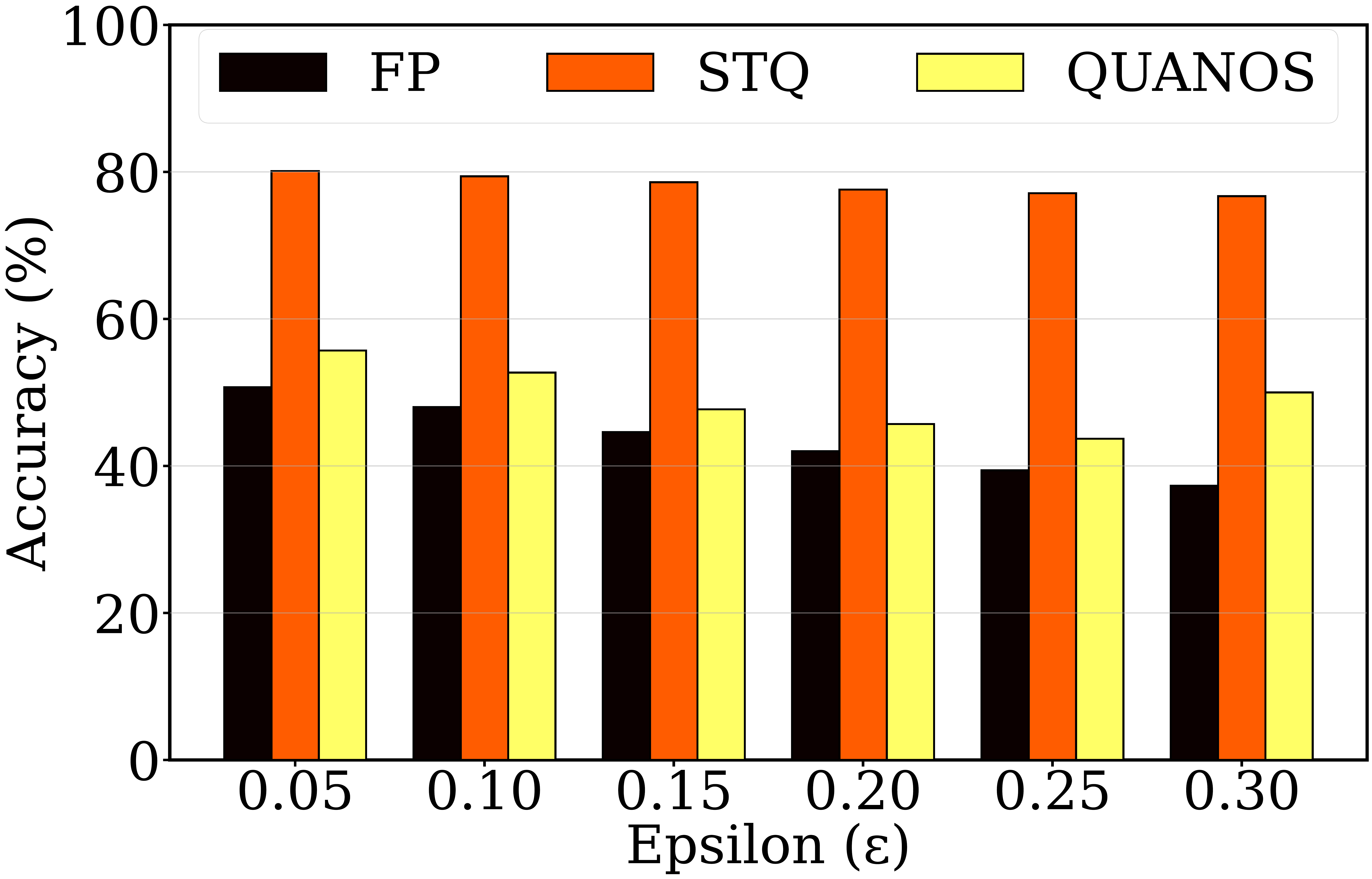}
    \caption{FGSM}
  \end{subfigure}
  \begin{subfigure}[b]{0.32\linewidth}
    \centering
    \includegraphics[width=0.99\linewidth]{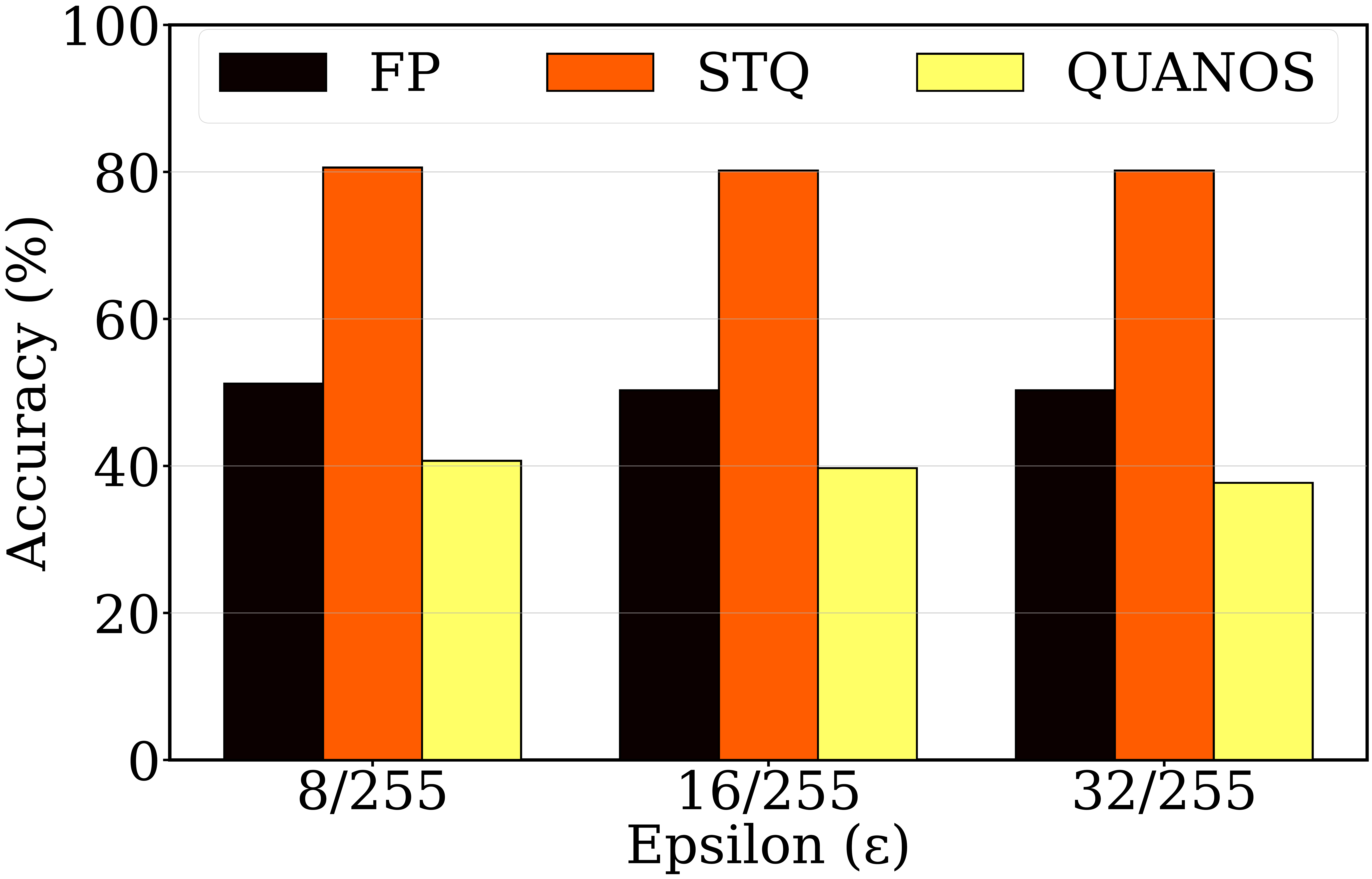}
    \caption{PGD}
  \end{subfigure}
  \begin{subfigure}[b]{0.32\linewidth}    
    \centering
    \includegraphics[width=0.99\linewidth]{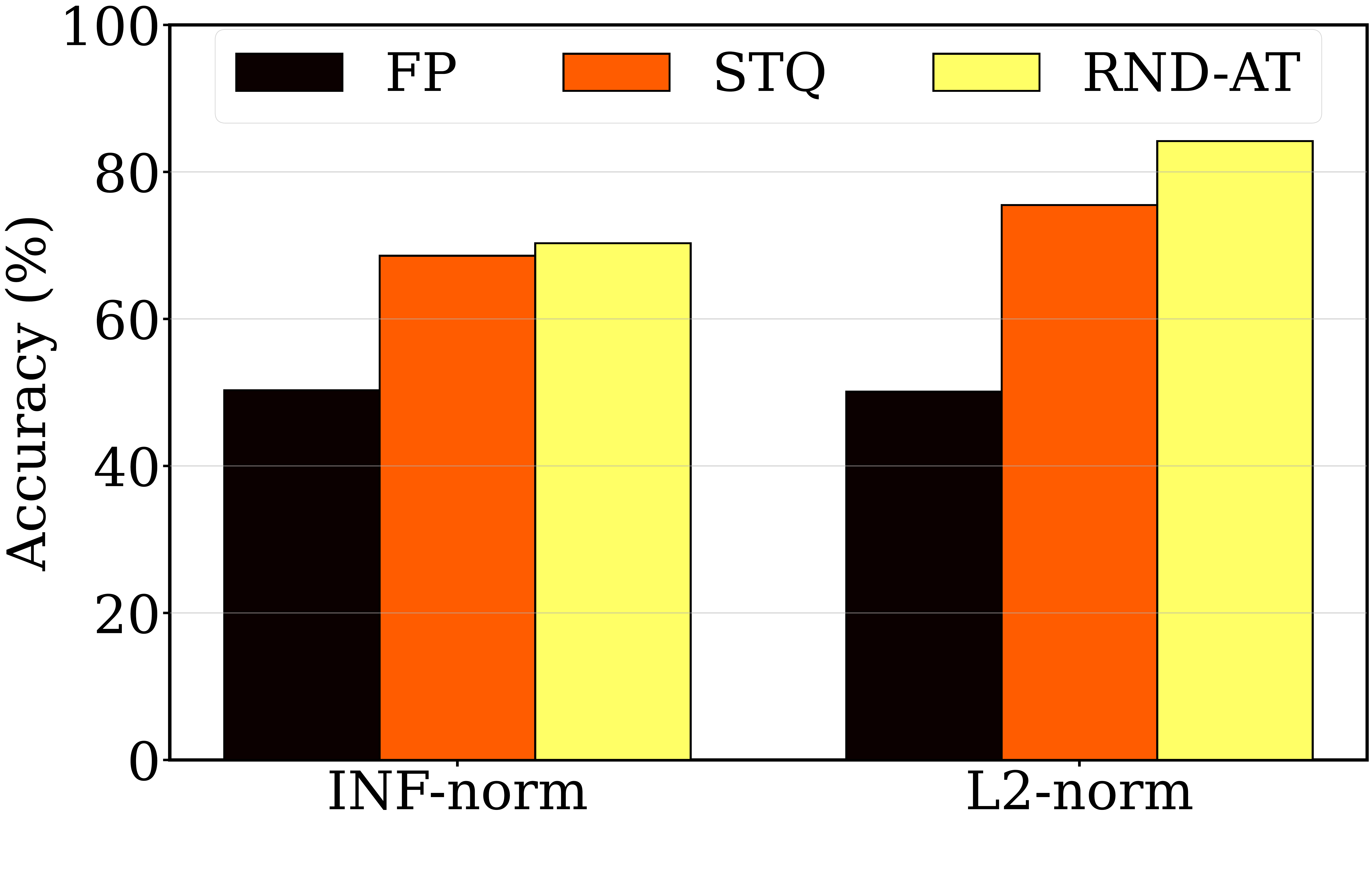}
    \caption{Square}
  \end{subfigure}
\caption{Robustness comparison of our FP and STQ models against Quanos~\cite{panda2020quanos} and RND-AT~\cite{qin2021random} for various FGSM and PGD perturbation strengths and variations of Square attacks.}
\label{fig:frameworks}
\end{figure*}

However, the accuracy of the proposed STQ model was either improved or slightly decreased in the presence of attacks. The largest accuracy drops of 7.0\% and 5.6\% were observed for the Square and C\&W-L2 attacks, respectively, and the CIFAR-10 dataset, although both outperformed the ResNetv1 benchmark. 
Note that ResNetv1 performs 0.9\% better than our STQ model only in the case of FGSM and PGD perturbations for the SVHN dataset. 
Finally, our model outperforms the DS-CNN model with better accuracy for both clean and adversarial samples of the GSC dataset, for all the experiments.

These results demonstrate that the resilience and robustness of the proposed STQ model against several adversarial attacks is better than the MLCommons/Tiny benchmarks tested models. As such, the per-layer JR integration of our STQ model has an advantage of reducing model complexity and in parallel enhancing the robustness of the model, producing an effective performance in comparison with traditional benchmark models. This is an interesting finding when looking at the requirements of MCUs, and other tiny devices that are memory and computational resource-constrained.

Table~\ref{table2} shows the variance reports of $K$-fold cross validation, where $K=10$. The MLC/T column represents ResNetv1 model for CIFAR-10 and SVHN datasets and DS-CNN model for the GSC dataset. The variance of the benchmark for the STQ model is lower in many samples of CIFAR-10 and all instances of the GSC dataset. For the SVHN, the variance of the ResnetV1 model is lower than that of STQ. 
Therefore, the $K$-fold cross validation results are varying with datasets and types of attacks, although the overall average results demonstrate general consistency of STQ, with categorical cross entropy loss. Particularly for the GSC dataset, at which STQ tends to outperform the MLC/T with both clean and attacks samples. These results demonstrate the performance capability of STQ as a robust and effective model suitable to be deployed into a resource-constrained environment.

\begin{table} [tp] 
\caption{Variance as a result of $K$-fold cross validation.}
\begin{center}
\label{table2}
\begin{tabular}{|c|c|c|c|c|c|c|}
\hline
\textbf{Dataset} & \textbf{Procedure} & \textbf{STQ} & \textbf{MLC/T} \\ \hline 
\multirow{5}{*}{CIFAR-10} &{No Attack} & \textbf{3.04} & {26.25}  \\ \hhline{~-||-----} 
& {FGSM} & \textbf{5.81} & {6.01}  \\ \hhline{~-||-----}  
& {PGD} & {3.7} & {3.35}  \\ \hhline{~-||-----} 
& {Square} & \textbf{11.78} & {15.58}  \\ \hhline{~-||-----}  
 & Boundary & \textbf{1.4} & {5.69}  \\ \hline 

\multirow{5}{*}{SVHN} &{No Attack} & {0.19} & {0.1}   \\ \hhline{~-||-----} 
& {FGSM} & {4.27} & {1.99}  \\ \hhline{~-||-----}  
& {PGD} & {1.64} & {1.40}   \\ \hhline{~-||-----}
& {Square} & \textbf{4.2} & {5.36}   \\ \hhline{~-||-----}  
 & Boundary & \textbf{0.39} & {0.45} \\ \hline 

 \multirow{5}{*}{GSC} &{No Attack} & \textbf{0.37} & {0.80}   \\ \hhline{~-||-----}  
& {FGSM} & \textbf{1.57} & {4.45} \\ \hhline{~-||-----}  
& {PGD} & \textbf{3.17} & {5.45}  \\ \hhline{~-||-----} 
& {Square} & \textbf{2.96} & {3.64}  \\ \hhline{~-||-----} 
 & Boundary & \textbf{0.24} & {0.34}  \\ \hline 
 & {Average} & \textbf{2.98} & {5.39}  \\ \hline 
\end{tabular}
\end{center}
\end{table}

Finally, Figure~\ref{fig:frameworks} shows a comparison between our FP and STQ models against: i) QUANOS~\cite{panda2020quanos} baseline model for various FGSM and PGD perturbation strengths; and ii) RND-AT~\cite{qin2021random} for variations of Square \cite{square} attacks. All attacks used 10,000 testing samples of the CIFAR-10 dataset. As we can see, STQ clearly outperforms QUANOS in detecting FGSM and PGD attacks for various strengths, and is slightly worse than RND-AT for Square attacks. We did not find any previous work for the Boundary attack which we could fairly compare against, and we leave Zoo and C\&W-L2 for future work.

\section{Conclusion} 
\label{conc}

This paper investigated the robustness of a Deeply Quantized Machine Learning (DQML) model against various white-box and black-box adversarial attacks. The deep quantization facilities of QKeras were considered to create a memory optimized, accurate and adversarially robust quantized model. This is due to its similarities with a defense technique, Jacobian Regularization (JR), that was integrated into it. 
As demonstrated, the proposed Stochastic Ternary Quantized (STQ) model, with quantization-aware training procedure introducing per-layer JR, was more robust than industry adopted MLCommons/Tiny benchmarks when facing several adversarial attacks. 
In fact, the stochastic quantization scheme was effective for model compression with support for robustness against adversarial attacks. This robustness was experimentally proved by observing its accuracy under various adversarial attacks utilizing two image (CIFAR-10 and SVHN) datasets and one audio (GSC) dataset. 
This is relevant in the context of deploying efficient and effective models in resource-constrained environments, with limited capabilities. Our initial results suggest further exploration of other sophisticated white-box and black-box attacks with different attack strengths. Future work will further assess the effectiveness of the proposed STQ QKeras model against new and latest attacks.


\section*{Acknowledgment}

This work has been supported by the PETRAS National Centre of Excellence for IoT Systems Cybersecurity, funded by the UK EPSRC under grant number EP/S035362/1.


\bibliographystyle{IEEEtran}
\bibliography{main}

\end{document}